\documentclass[runningheads]{llncs}
\usepackage{graphicx}
\usepackage{amsmath}
\usepackage{booktabs}
\usepackage{amsmath,amssymb,amsfonts}
\usepackage{graphicx}
\usepackage{algorithm}
\usepackage{algorithmic}

\usepackage{tikz}
\usepackage{textcomp}
\usepackage{hyperref}
\usepackage{lipsum}

\newcommand\copyrighttext{%
  \footnotesize \footnotesize Final version published in the proceedings of the 29th International Conference on Artificial Neural Networks, ICANN 2020. The final authenticated publication is available online at \url{https://doi.org/10.1007/978-3-030-61609-0_52}}
\newcommand\copyrightnotice{%
\begin{tikzpicture}[remember picture,overlay]
\node[anchor=south,yshift=10pt] at (current page.south) {\fbox{\parbox{\dimexpr\textwidth-\fboxsep-\fboxrule\relax}{\copyrighttext}}};
\end{tikzpicture}%
}

\begin{document}
\title{Tell Me Why You Feel That Way: \\ Processing Compositional Dependency
for \\Tree-LSTM Aspect Sentiment Triplet Extraction (TASTE)}
\titlerunning{Tree-LSTM Aspect Sentiment Triplet Extraction}
\author{Alexander Sutherland\inst{1} \and
Suna Bensch\inst{2} \and
Thomas Hellström\inst{2} \and
Sven Magg\inst{1} \and
Stefan Wermter\inst{1}}

\authorrunning{A. Sutherland \textit{et al.}}
\institute{
Department of Informatics, University of Hamburg \\
Vogt-Koelln-Str. 30, 22527 Hamburg, Germany \\
\email{\{sutherland,magg,wermter\}@informatik.uni-hamburg.de}\\
%\url{http://www.springer.com/gp/computer-science/lncs} 
\and
Department of Computing Science, Umeå University, Umeå, Sweden\\
\email{\{suna,thomash\}@cs.umu.se}}
\maketitle              % typeset the header of the contribution
\copyrightnotice
\begin{abstract}
Sentiment analysis has transitioned from classifying the sentiment of an entire sentence to providing the contextual information of what targets exist in a sentence, what sentiment the individual targets have, and what the causal words responsible for that sentiment are. However, this has led to elaborate requirements being placed on the datasets needed to train neural networks on the joint triplet task of determining an entity, its sentiment, and the causal words for that sentiment. Requiring this kind of data for training systems is problematic, as they suffer from stacking subjective annotations and domain over-fitting leading to poor model generalisation when applied in new contexts. These problems are also likely to be compounded as we attempt to jointly determine additional contextual elements in the future. To mitigate these problems, we present a hybrid neural-symbolic method utilising a Dependency Tree-LSTM's compositional sentiment parse structure and complementary symbolic rules to correctly extract target-sentiment-cause triplets from sentences without the need for triplet training data. We show that this method has the potential to perform in line with state-of-the-art approaches while also simplifying the data required and providing a degree of interpretability through the Tree-LSTM.

\end{abstract}
\section{Introduction}
Sentiment Analysis (SA) has been described as a suitcase \cite{cambria2017sentiment}, where SA is composed of several smaller Natural Language Processing (NLP) tasks that need to be performed to acquire a contextual understanding of a predicted sentiment label. One of the most prevalent steps beyond sentence-level sentiment classification is Aspect-Based Sentiment Analysis (ABSA) \cite{pontiki2016semeval}. In ABSA, targets in sentences for a particular domain are determined and assigned specific sentiment labels that may differ from other targets in the same sentence. Jointly being able to determine a target and its sentiment is a necessary step towards being able to use that information in practical scenarios.

%Contextual information is important for intelligent agents, such as virtual assistants or robots. These agents exist within an environment they have to interpret and therefore contextual information is vital to determine what sentiment information should be applied or ignored given the agent's task at hand. While agents are still able to make high-level decisions based on the state of the current interaction \cite{siqueira2018disambiguating}, information such as the targets in an affective sentence provide agents with far more information that can then be analysed to adjust the decision-making process. However, even more information other than targets and their sentiment is sometimes required to make useful decisions.

Extracting terms used to identify the cause or reason for a sentiment outcome in a sentence is sometimes called Opinion Term Extraction (OTE) \cite{peng2019knowing}. The goal of OTE is to identify these causal words in a sentence and compare them to user annotations of what they consider to be the cause of a sentiment. OTE provides an idea of what words convey affective meaning. In recent works, ABSA and OTE have been combined \cite{peng2019knowing} to create triplets of information describing a target, its sentiment, and the cause of that sentiment. This task, known as Aspect Sentiment Triplet Extraction (ASTE) \cite{peng2019knowing}, has the potential to provide a lot of contextual information to intelligent agents allowing them to not only know what is being spoken about but how an individual feels about it and what the cause of that feeling may be. With this information, the agent can either reinforce a certain behaviour based on the cause, if positive, or attempt to remedy a negative cause if possible. An example of such a triplet for the sentence ``the food was excellent'' would be (``the food'', positive, ``excellent'') where ``the food'' is the target, the sentiment is positive, and ``excellent'' is the opinion term.

While predicting sentiment triplets is valuable, the method in which we acquire them for agents needs to be robust. Agents often find themselves in dynamic contexts provided by natural language when interacting with individuals in virtual and real-world environments. These context shifts are very problematic when doing triplet sentiment analysis, as the new targets or aspects that occur may not have been captured by the training data. Likewise, the process of having three subjective labels that have to be jointly predicted is difficult to learn and difficult to expand upon, as any new datasets would require annotators to make three subjective decisions regarding what substring constitutes a target, what the sentiment of that target is, and what the cause for that sentiment is. 

These tasks have the possible flaw of introducing annotator bias and the alignment of triplet information introduces further bias. As an example, the original opinion terms were aligned with aspects by only two annotators \cite{fan2019target} and we do not know how many annotators the opinion terms had. Across all of the papers used by the triplet dataset \cite{peng2019knowing} this has resulted in 5 different degrees of subjective error for the individual tasks and alignments combined during training and evaluation. While these errors could be mitigated by naively relabelling the data set with triplets from scratch, it would not solve shortcomings in scenarios in which we would want additional context as quadruplet or more or would like to change into a new domain. Methods that require specific jointly annotated data do not scale well between domains. Therefore we propose leveraging a combination of symbolic information and neural processing to circumvent this.

As an alternative, we propose a hybrid neural-symbolic method for Tree Long-Short Term Memory (Tree-LSTM) Aspect Sentiment Triplet Extraction (TASTE) through the symbolic analysis of the compositional processing of a Dependency Tree-LSTM (DTLSTM) \cite{tai2015improved}. This approach has the benefit of not requiring domain-specific triplet training data but only requires sentiment annotations from the Stanford Sentiment Treebank (SST) \cite{socher2013recursive}. This method scales between domains as it only relies on the compositional processing done by the DTLSTM and noun chunk identification as opposed to learning high-level relations at domain-level between labels. Furthermore, due to the structured recursive processing of the DTLSTM, we are granted a degree of interpretability through predictions over the dependency parse tree, allowing for common-sense reasoning as to why a certain target, sentiment, and causes were selected. We show that this method allows us to perform in line with state-of-the-art neural approaches under the correct circumstances while allowing for the aforementioned interpretability and simple output and behaviour modifications through symbolic rules.

\section{Related Work}
%Methods of being able to identify the dynamics of sentiment and semantics in language are still developing. \cite{socher2013recursive} develop a method of recursively processing language based on the syntactic structure of a sentence to show the sentiment prediction for every node in a syntax tree. This was later extended by \cite{tai2015improved} who incorporate this design into the Tree-LSTM to utilise the benefits of the LSTM \cite{hochreiter1997long} and in particular introduce the Child-sum Tree-LSTM which can process an arbitrary number of child nodes. In particular, when this method is applied to a dependency parse tree it is called a DTLSTM. 

Sutherland \textit{et al.} \cite{sutherlandleveraging} provide a method of identifying targets and their sentiment based on the sentiment parse of a Tree-LSTM over a dependency parse tree. This is done by applying a set of symbolic rules to the neurally determined output of the Tree-LSTM for every node in the syntactic dependency parse tree, where each prediction is based on hidden states of the children of a node. This method relies upon the structure of dependency parses having verbs as the heads of affective substrings that contain targets, wherein the targets are children of the verb. In this paper, we extend the work of Sutherland \textit{et al.} to identify more targets and include cause, also known as Opinion Term Extraction (OTE).

Using dependency trees is beneficial for ABSA and often involves symbolic operations on the dependencies. Wang \textit{et al.} \cite{wang2016recursive} present Recursive Neural Condition Random Fields (RNCRF) which utilise a recursive network to produce high-level representations based on the syntactic structure. They are then used to train conditional random fields to extract aspect-opinion pairs. Dai \textit{et al.} \cite{dai2019neural} also trained a Bi-directional LSTM-Conditional Random Field (BiLSTM-CRF) which utilises additional unlabelled data together with the SemEval data \cite{pontiki2016semeval} to determine aspect-opinion pairs. 

Peng \textit{et al.} \cite{peng2019knowing} go so far as to actively construct the triplet dataset and use a two-stage model of first predicting the target and opinion together, and then use a Bidirectional LSTM to predict if the two are a valid pair in the second stage. This allows the system to utilise information from the opinion extraction when predicting targets to get a higher performance overall. In our work, we utilise the data provided by Peng \textit{et al.} and symbolic rule-based principles from Sutherland \textit{et al.} \cite{sutherlandleveraging} to show that the task of ASTE can be performed comparatively under the right conditions with only access to standard sentiment data.

\section{SemEval ASTE Dataset}
We utilise the SemEval ASTE dataset \cite{peng2019knowing} which is an extension of the SemEval ABSA dataset \cite{pontiki2016semeval}. The SemEval ABSA task, proposed at the 2014 International Workshop on Semantic Evaluation (SemEval), is based on identifying targets of sentiment in sentences. The SemEval ABSA dataset was extended by Pontiki \textit{et al.} \cite{pontiki2016semeval} to include additional elements such as being able to identify the particular category a target may belong to or being able to perform ABSA in different languages. This dataset was used as the foundation to further deepen the analysis of sentiment by introducing \textit{``opinion terms''} \cite{wang2016recursive}, what can be considered to be the reason or cause for a particular sentiment. The alignments between the aspects and the opinion terms are provided by Fan \textit{et al.} \cite{fan2019target}.

This all culminated in the work of Peng \textit{et al.} \cite{peng2019knowing}, wherein they introduce the SemEval ASTE dataset\footnote{https://github.com/xuuuluuu/SemEval-Triplet-data}, consisting of identifying targets, classifying the sentiment towards those targets, and determining which words in a sentence were responsible for the sentiment towards a target, the \textit{``opinion''}. We use this extended dataset, which has been provided by the authors to analyse how well our approach can extract sentiment triplets and to compare against other approaches. Each data-point consists of a sentence and each sentence can have multiple targets which are substrings in the sentence. Each target has a specific sentiment, with the labels positive, neutral, and negative, and an opinion term which consists of a substring in the sentence defined by annotators. 

The dataset itself is split up into two domains restaurants and laptops: there are three versions of the restaurant dataset, denoted rest14, rest15, and rest16 based on the year the data was used in the SemEval challenge, and lap14 for the 2014 laptop ABSA data. The total number of viable targets for each version can be seen in Table \ref{PRF}.

\section{Dependency Tree-LSTM}
Our method is based on the compositional processing used by the Dependency Tree-LSTM (DTLSTM) \cite{tai2015improved}, that was utilised for ABSA by Sutherland \textit{et al.} \cite{sutherlandleveraging}, whose method functions as a basis for our approach. To extract our sentiment triplet we need to know the distribution of sentiment over a dependency tree. To do this, we pre-train a DTLSTM on the Stanford Sentiment Treebank (SST) \cite{socher2013recursive} training data. What differentiates a DTLSTM and an LSTM is that it can take several inputs at each time-step and processes input based on the dependency tree parse structure of its input sentence. A DTLSTM has the weight matrices $W$ and $U$, and a bias vector $b$ similar to the LTSM it is based upon with the same sigmoidal activation function $\sigma_g$. 

For a DTLSTM, a node $j$ in a dependency tree possesses a set of dependent child nodes, denoted $C(j)$, which contains the indexes of children $k$, with each child possessing a hidden state $h_k$. An input word vector $x_j$ is the vector corresponding to the word associated with $j$ in the parse tree. The summed hidden states $\tilde{h}_{j}$ of the child nodes are the input to the input gate $i_j$. The DTLSTM has several forget vectors $f_{jk}$ for each child $k$ of node $j$. The DTLSTM has an output gate activation vector $o_j$, such as those possessed by an LSTM, and a memory cell vector $u_j$. The updated memory cell state vector $c_j$ is the summed element-wise multiplication of $i_j$ and $u_j$ with the summed element-wise multiplications of each child cell state $c_k$ with $f_{jk}$. The hidden state $h_j$ is the element-wise multiplication of $o_j$ and $tanh(c_j)$.  The equations for the DTLSTM, are defined by Tai \textit{et al}. \cite{tai2015improved} and are as follows:

\begin{align}
& \tilde{h}_{j} = \sum_{k \in C(j)} h_k, \\
& i_j = \sigma_g(W^{(i)}x_j + U^{(i)} \tilde{h}_{j} + b^{(i)}), \\
& f_{jk} = \sigma_g(W^{(f)}x_j + U^{(f)} {h}_{k} + b^{(f)}), \\
& o_{j} = \sigma_g(W^{(o)}x_j + U^{(o)} \tilde{h}_{j} + b^{(o)}), \\
& u_{j} = \sigma_g(W^{(u)}x_j + U^{(u)} \tilde{h}_{j} + b^{(u)}), \\
& {c}_{j} = i_j \odot u_j + \sum_{k \in C(j)} f_{jk} \odot c_k, \\
& {h}_{j} = o_j \odot tanh(c_j).
\end{align}

Words in a sentence are represented as Word Embeddings from the pre-trained Common-Crawl 840B data\footnote{https://nlp.stanford.edu/projects/glove/} before they are fed to the DTLSTM. To analyse sentiment for sentences according to a tree structure, we use the state of the art NLP library SpaCy \cite{honnibal2017spacy} to extract dependency trees. We see when predicting over dependency trees that the sentiment that propagates through the tree can come from many different sources, such as nouns, adjectives, and sometimes the verb heads themselves \cite{socher2013recursive}. The compositional structure also allows for the extraction of sub-trees as opposed to other approaches that only provide individual words. 

\section{Aspect Sentiment Triplet Extraction}
To analyse how close the features utilised by our neural models are to the annotations provided by a user we have to determine what are viewed as salient features. We do this by observing two TASTE approaches that can allocate salience to words in a sentence when considering the sentiment. 

\begin{figure*}[!htb]
\centering
  \includegraphics[width=\linewidth]{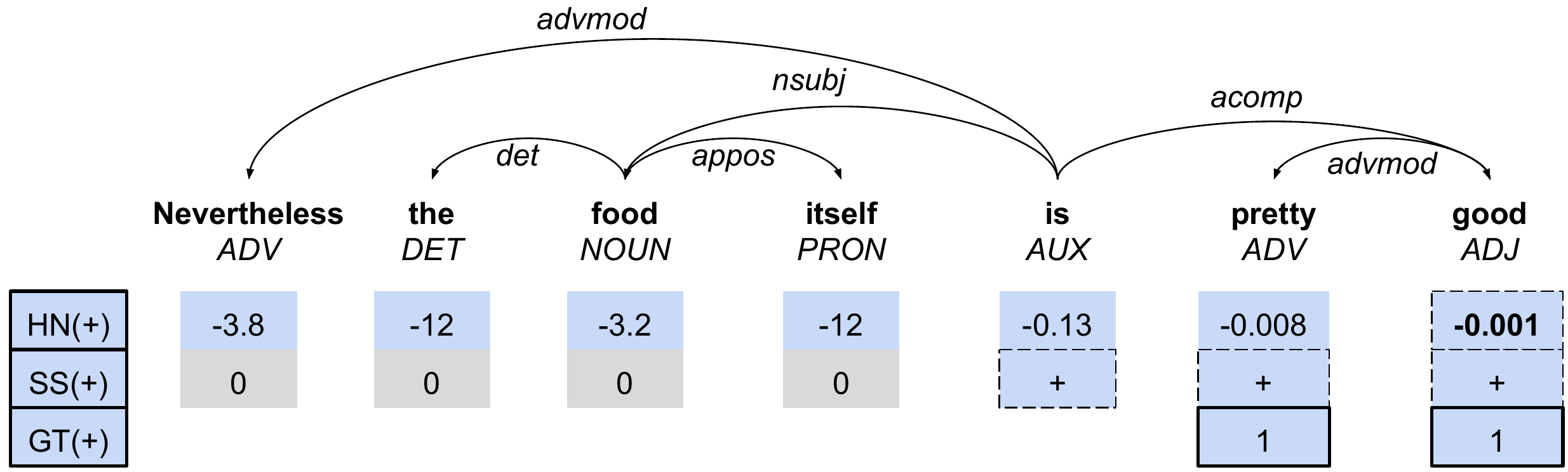}
  \caption{An example of what kind of sentiment opinion terms are identified from a sentiment dependency parse for the positive sentiment (+) in an example sentence with the target ``food'' and the sentiment ``positive'' for our two methods: Highest Node (HN) and Sentiment Search (SS). The selected reason candidates for the respective methods, as represented by the dashed boxes, can be compared against the ground truth reasoning (GT) as provided by annotators. In this sentence, we see that HN identifies the word ``good'' as the reason, whereas SS identifies the substring ``is pretty good'', and annotators identified ``pretty good'' as the correct reason. The values in the HN row are the logarithmic softmax values for the positive sentiment class of the projection layer of the Tree-LSTM.}
  \label{fig:parse1}
\end{figure*}

\subsection{Target identification}
\label{ti}

Our method of identifying affective targets extends the method of Sutherland \textit{et al.} \cite{sutherlandleveraging} by changing the conditions under which targets are extracted. In the work of Sutherland \textit{et al.}, targets are extracted when there exists a noun chunk, as identified by SpaCy, that has a verb as its head in the dependency tree. This provided both a target and the sub-tree it belongs to as defined by the children of the head verb.

We adjust this method by addressing cases where there is no verb or the verb is defined as an auxiliary verb, and therefore possesses the \textit{``AUX''} dependency tag. The same basic algorithm is followed: for every verb or auxiliary verb in a sentence, if a noun chunk exists as a substring within the substring created from that parent's children and itself then that noun chunk is a viable target. Every target can only be associated with its closest parental verb but a parental verb may possess multiple noun chunks. In addition to this, we allow dependency trees that have noun chunks but no verbs to also be considered as targets. An example sentence where this would be relevant would be the sentence \textit{``Good food.''}, where \textit{``food''} would become the target.

\subsection{Target sentiment classification}

Sutherland \textit{et al.} \cite{sutherlandleveraging} leveraged a method classifying the sentiment of a target by having the target inherit the sentiment of its parental verb in a trickle-down fashion. This is possible thanks to the DTLSM's compositional parsing and the fact that sentiment is propagated up through the dependency tree to the root of the sentence. We utilise the same method of sentiment prediction, by using the DTLSTM to predict a sentiment vector over the negative, neutral, and positive classes for every node in the dependency parse tree.

The DTLSTM is pre-trained on the SST \cite{socher2013recursive} and retains the hyper-parameter set as reported by Tai \textit{et al.} \cite{tai2015improved}. We use these hyper-parameters as the default in the implementation we adjusted for our approach\footnote{https://github.com/ttpro1995/TreeLSTMSentiment}, as we found adjustments to be insignificant and further optimisation of the architecture is outside of the scope of this paper. Similar to Sutherland \textit{et al.} \cite{sutherlandleveraging} and Peng \textit{et al.} \cite{peng2019knowing}, we select the pre-trained model that performed the best on the SST validation data over ten epochs. With this, we can acquire a prediction for every sentiment class for every node in the dependency tree. 

Once this is done we iterate over every target and each target inherits the sentiment of its parental verb, essentially trickling down the sentiment from the verb to the noun chunks. As an example, sentence \textit{``the food is good''} would have a positive sentiment propagated from the word good up to the root \textit{``is''}. As the target \textit{``the food''} has \textit{``is''} as its parental verb, it will also inherit the positive sentiment that has been propagated up to \textit{``is''}. In the case of sentences or sub-sentences where there are no verbs, the noun of the noun chunk acts as the root of the sentence and is the receiver of the sentiment charge. %Therefore no inheritance is required.

\subsection{Target opinion term extraction}

We examine two different TASTE approaches for extracting cause from a sentiment dependency parse. The first approach, Highest Node (HN), is to recursively search the predictions of the sub-tree of a target from the parental node of the target and extract the node in its sub-tree with the highest logarithmic softmax activation level for the intended sentiment. This provides us with a single word from the dependency parse tree that is believed to be the most likely reason for the outcome sentiment. The benefit of this approach is that it is likely to reliably identify at least one word in the reason, as additional non-salient words are likely to reduce the activation level as opposed to increase it. Drawbacks are that reasons that require multiple words to understand, e.g. \textit{``very good''} cannot be fully extracted. Another issue is that if a parental node is somewhat trending toward the sentiment, but not the primary cause, it may be picked instead.

The second approach, Sentiment Search (SS), allows for the extraction of this additional causation context, as we recursively search through the sub-tree of the parent node of a target and extract all nodes where the target sentiment is higher than the other classes. This allows for the selection of multiple reason words and the construction of multi-word substrings. So for the sentence \textit{``The sushi was pretty good''} \textit{``was''}, \textit{``pretty''} and \textit{``good''} would be extracted as reasons for the positive sentiment towards sushi. 

The drawbacks of this method are that it can suffer from low precision, as it will pick up words that inherit positive sentiment as they propagate up the dependency tree, and that it cannot identify a reason if positive sentiment is not being properly propagated up the tree, i.e. nodes that incorrectly have a higher sentiment for neutral rather than positive. Examples of what both methods identify and the kind of features they look at can be seen in Figure \ref{fig:parse1}. The method used for recursively going through the affective dependency parse tree and picking out either HN or SS is defined in Algorithm~\ref{alg1}, which recursively searches through the tree of nodes from the root in a depth first fashion to identify the SS and HN nodes.

\begin{algorithm} % enter the algorithm environment
\caption{RecursiveSearch} % give the algorithm a caption
\label{alg1} % and a label for \ref{} commands later in the document
\begin{algorithmic}[1] % enter the algorithmic environment
\IF {$node_iSentiment $ \textbf{equals} $ targetSentiment$}
    \STATE {$SSHypothesis.insert(nodeWord)$}
\ENDIF
\IF {$len(children(node_i)) $ \textbf{is} $ 0$}
    \STATE \textbf{return} $node_iSentActivation$
\ELSE
    \STATE {$highestActivation \gets node_iSentActivation$}
    \STATE {$HNHypothesis \gets nodeWord$}
    
    \FOR{$c$ \textbf{in} $children(node_i)$}
        \STATE {$temp \gets RecursiveSearch(node_i)$}
        \IF{$temp \geq highestActivation$}
            \STATE {$highestActivation \gets temp$}
            \STATE {$HNHypothesis \gets childWord(c)$}
        \ENDIF
        
    \ENDFOR
    \STATE \textbf{return} $highestActivation$
\ENDIF

\end{algorithmic}
\end{algorithm}

\section{Evaluation}
Google's Bilingual Evaluation Understudy (Google-BLEU or GLEU) \cite{wu2016google} is used to measure distance between our extracted targets and what are designated as such in the dataset in addition to precision and recall. We choose to include GLEU as one of our measurements at extraction level, as we wish to take a more inclusive stance on what is considered to be correctly extracting a target or opinion term. For example, the sentence \textit{``I loved the red cake''} may have a given target \textit{``cake''}, however, the \textit{``red cake''} or even \textit{``the red cake''} are also equally valid targets for the positive sentiment in that sentence. 

As such we consider a target or opinion term extracted if they are equal to or exist as a substring of the annotation or vice-versa. Therefore, we provide the GLEU to give an estimate on how far our extraction strays from the original label, as we believe collecting additional contextual information for a target is beneficial and that this metric also addresses that the annotations are subjective and not always objectively correct. When calculating GLEU for targets and SS we remove determiners and copula, as annotators did not deem these as part of the labels and removing them does not change the semantic meaning. For determiners we remove ``the'', ``a'', and ``an''. For copula we remove ``is'', ``was'', ``were'', and ``are''.

GLEU, as defined by Wu \textit{et al.} \cite{wu2016google} and as implemented in the NLTK python package \cite{perkins2014python}, is calculated for a generated and reference sentence by calculating recall and precision for all sub-sequences of 1 to 4 tokens between the two sentences, where recall is the ratio of correct sub-sequences in the reference sentence and precision is the ratio of matching sub-sequences in the generated sentence. The GLEU score itself is the minimum of recall and precision, selecting the lowest value, with the lowest value being 0, indicating no matching sub-sequences and the highest possible value being 1, indicating a full match between all possible sub-sequences for both precision and recall. In this section, we present the cascading results of target identification, sentiment classification for that target, and the extraction of the opinion term for that target given its sentiment.

\subsection{Target identification results}
In Table \ref{PRF} we present the results of target identification from sentences using the symbolic method specified in Section \ref{ti}. We see that the laptop dataset is more difficult to extract targets from when compared to the restaurant datasets. 

\begin{table}[H]
\centering
\caption{Targets, precision, and recall for the identified targets over all of the datasets as provided by our symbolic extraction method. In addition to this, the average GLEU is provided to indicate how well the identified target tokens match the labelled ground truth tokens. A reference for the strength of the GLEU can be found in Table \ref{GLEU-results}, where we show GLEU for labels against the entire sentence and single words.}
\begin{tabular}{|l|c|c|c|c|}
\hline
 & Targets & P & R & Avg. GLEU \\ \hline % With dets
14res & 849 & 0.549 & 0.921 & 0.722 \\ \hline %0.430
14lap & 475 & 0.426 & 0.785 & 0.713 \\ \hline %0.509
15res & 426 & 0.461 & 0.920 & 0.731 \\ \hline %0.492
16res & 444 & 0.476 & 0.921 & 0.714 \\ \hline %0.482
\end{tabular}
\label{PRF}
\end{table}

\subsection{Targeted sentiment classification results}

Here we present the results of the targeted sentiment analysis after target identification, which can be seen in Table \ref{Sent-acc}. In general, we see that sentiment performance between restaurants and laptops is fairly uniform, indicating that the sentiment propagation and reliance on parental verbs for classification does not differ with as large a margin as for the tasks of target identification and OTE. 

\begin{table}[H]
\centering
\caption{Overall sentiment accuracy for target sentiments over all of the datasets based on the targets that were identified.}
\begin{tabular}{|l|c|c|c|}
\hline
 & Accuracy \\ \hline 
14res & 0.740  \\ \hline 
14lap & 0.719 \\ \hline 
15res & 0.722 \\ \hline 
16res & 0.778  \\ \hline 
\end{tabular}
\label{Sent-acc}
\end{table}

\subsection{Targeted opinion term extraction results}
In this section, we show the results of OTE and triplet extraction. In Table \ref{results} we see the results of applying \textbf{HN}, \textbf{SS} and merging the set of extracted terms for \textbf{HN \& SS} for OTE.  We also compare these results against GLEU for the entire sentence (\textbf{FULL}) to show that the extracted data is more semantically relevant. We see that SS can extract more opinion terms than HN, however, we also see in Table \ref{GLEU-results} that HN tends to give predictions closer to what is labelled by annotators. Interestingly, we see in Table \ref{results} that there is a misalignment between extracted opinions and sentiment, otherwise, both rows would be mirrored by their ``-3'' counterparts. This implies that the system is extracting the correct target despite using the incorrect sentiment when recursively searching.

\begin{table}[H]
\centering
\caption{F1-scores of the HN and SS methods for extracting opinion terms for the targets found in Table \ref{PRF}. Rows denoted with ``-3'' refer to when the sentiment and opinion term are correctly identified for a correctly extracted target in the previous step, thus successfully identifying a full triplet. As a triplet is generated for every found target, the F1-score is equivalent to the precision and recall under these conditions.}
\begin{tabular}{|l|c|c|c|c|}
\hline
 & 14res & 14lap & 15res & 16res \\ \hline
 HN & 0.482 & 0.398 & 0.515 & 0.503 \\ \hline
 SS & \textbf{0.625} & \textbf{0.488} & \textbf{0.633} & \textbf{0.615} \\ \hline \hline
 HN-3 & 0.420 & 0.315 & 0.450 & 0.447 \\ \hline
 SS-3 & \textbf{0.547} & \textbf{0.386} & \textbf{0.547} & \textbf{0.543} \\ \hline
\end{tabular} \label{results}
\end{table}

\begin{table}[H]
\centering
\caption{GLEU scores at opinion level for our two different opinion targeting identification methods with the copula ``is'', ``was'', ``s'', ``were'', and ``are'' removed for the SS method. We include the full sentence score against the label to provide a contrast to the selection methods.}
\begin{tabular}{|l|c|c|c|c|c|c|c|c|}
\hline
 & \multicolumn{2}{c|}{14res} & \multicolumn{2}{c|}{14lap} & \multicolumn{2}{c|}{15res} & \multicolumn{2}{c|}{16res} \\ \hline
 & Avg. GLEU & $\sigma$ & Avg. GLEU & $\sigma$ & Avg. GLEU & $\sigma$ & Avg. GLEU & $\sigma$ \\ \hline
Full Sent & 0.036 & 0.049 & 0.041 & 0.050 & 0.056 & 0.079 & 0.056 & 0.075 \\ \hline
HN & \textbf{0.797} & 0.342 & \textbf{0.736} & 0.376 & \textbf{0.786} & 0.344 & \textbf{0.778} & 0.349 \\ \hline
SS & 0.377 & 0.381 & 0.377 & 0.377 & 0.411 & 0.362 & 0.398 & 0.377 \\ \hline
%SS GLEU & 0.195 & 0.168 & 0.264 & 0.293 & 0.248 & 0.223 & 0.248 & 0.226 \\ \hline % GLEU with Copula
\end{tabular} \label{GLEU-results}
\end{table}

\begin{table*}[t]
\centering
\caption{Precision and recall scores for the correctly predicted triplets from each version of the restaurant and laptop dataset for our method against different benchmarks. TD stands for triplet data and denotes how much triplet training data is required to achieve the specified results. Methods with a ``+'' are the cascading results of models presented in the paper by Peng \textit{et al.} \cite{peng2019knowing}. TASTE* SS shows results of our SS method for cases with correct targets before triplet alignment based on Table \ref{results}. }

\begin{tabular}{|l|c c c|c c c|c c c|c c c|}
\hline
 & \multicolumn{3}{c|}{14res} & \multicolumn{3}{c|}{14lap} & \multicolumn{3}{c|}{15res} & \multicolumn{3}{c|}{16res} \\ \hline
 & TD & P & R & TD & P & R & TD & P & R & TD & P & R \\ \hline
RINANTE+  & 2669 & 0.376 & 0.340 & 1602 & 0.231 & 0.176 & 1161 & 0.294 & 0.269 & 1605 & 0.271 & 0.205  \\ 
CMLA+  & 2669 & 0.401 & 0.466 & 1602 & 0.314 & 0.346 & 1161 & 0.344 & 0.376 & 1605 & 0.436 & 0.398  \\ 
Peng \textit{et al.} \cite{peng2019knowing}  & 2669 & \textbf{0.442} & \textbf{0.629} & 1602 & \textbf{0.404} & \textbf{0.472} & 1161 & \textbf{0.410} & 0.547 & 1605 & \textbf{0.468} & \textbf{0.630}  \\ \hline
TASTE HN & \textbf{0} & 0.225 & 0.379 & \textbf{0} & 0.128 & 0.248 & \textbf{0} & 0.203 & 0.416 & \textbf{0} & 0.208 & 0.417\\ 
TASTE SS  & \textbf{0} & 0.293 & 0.493 & \textbf{0} & 0.157 & 0.304 & \textbf{0} & 0.246 & 0.505 & \textbf{0} & 0.253 & 0.507\\ 
TASTE HN \& SS  & \textbf{0} & 0.323 & 0.543 & \textbf{0} & 0.182 & 0.353 & \textbf{0} & 0.269 & \textbf{0.552} & \textbf{0} & 0.273 & 0.549\\ \hline
TASTE* SS  & \textbf{0} & \textbf{0.547} & 0.547 & \textbf{0} & 0.386 & 0.386 & \textbf{0} & \textbf{0.547} & 0.547 & \textbf{0} & \textbf{0.543} & 0.543\\ \hline

\end{tabular} \label{comparison}
\end{table*}

In Table \ref{comparison}, we see that recall values for SS and when HN \& SS are in line with state of the art approaches, even beating them for 15rest, in-spite of our method only using one pre-trained model and not being trained on triplet data. However, our model is penalised for generating more targets than what are considered aspects of that domain by annotators. If our approach was not being penalised for the out-of-domain targets, then our recall would become our F1-score.

\section{Discussion \& Conclusion}
We show in Table \ref{comparison} that we can perform ASTE without the need for triplet data if we can be flexible with acquiring targets and opinion terms. This is at the cost of precision, as if we require an exact GLEU match with the annotations then the average accuracies shown in Table \ref{results} will decrease, with HN dropping from 0.474 to 0.333 and SS more sharply declining from 0.590 to 0.151 as we approach a GLEU value of 1. The decline of SS being due to SS over-selecting tokens while HN will under-select tokens. However, in practice, the content will be largely the same, as the intended label still exists as a substring in the extracted opinion.

By interpreting DTLSTM output, our research aligns with recent work in interpretable AI for making decisions in AI systems understandable to humans \cite{HellBen18}. Current algorithms can determine some sentiment but methods for detecting what the cause of a sentiment is or who is targeted by the sentiment are not thoroughly investigated. The application areas of interpretable AI are vast including being able to explain the reasons for recruitment decisions, credit check decisions or to mitigate bias but also for intelligent agents or robots to successfully function alongside humans. For instance, robots that only detect an unhappy human cannot act without knowing the cause of this unhappiness.

We present a method of extracting target-sentiment-cause triplets, TASTE, without the need for domain-specific or triplet training data. We show that this method can retrieve as many correct triplets as state-of-the-art methods that require prohibitive triplet data. We believe that this method is useful for intelligent agents, such as robots, that would have to be able to interpret contextual sentiment information on the fly in a dynamic context and domain. 

% This applies under the condition that freedom can be taken with the preciseness with which aspect and opinion terms are extracted.

%Cases where agents would learn aspect terms limited to a single domain would cause the agents to be unable to find targets if they transition into a too different domain. 
Our approach can determine triplets regardless of how the context shifts, as targets are syntax-based and sentiment is learned compositionally, systems then having a selection of valid triplets that can be interpreted in a scenario. Furthermore, our system does not require the constant acquisition of new domains that require further triplet annotation collection, as our method learns hierarchical sentiment composition from the SST through the DTLSTM. In summary, our method allows for more domain independent ASTE that works in more dynamic contexts at the cost of precision.

\section*{Acknowledgements}
This work has received funding from the European Union's Horizon 2020 research and innovation programme under the Marie Sk\l{}odowska-Curie grant agreement No 721619 for the SOCRATES project.

\bibliographystyle{splncs04}
\bibliography{icann2020}

\begin{thebibliography}{10}
\providecommand{\url}[1]{\texttt{#1}}
\providecommand{\urlprefix}{URL }
\providecommand{\doi}[1]{https://doi.org/#1}

\bibitem{cambria2017sentiment}
Cambria, E., Poria, S., Gelbukh, A., Thelwall, M.: Sentiment analysis is a big
  suitcase. IEEE Intelligent Systems  \textbf{32}(6),  74--80 (2017)

\bibitem{dai2019neural}
Dai, H., Song, Y.: Neural aspect and opinion term extraction with mined rules
  as weak supervision. In: Proceedings of the 57th Annual Meeting of the
  Association for Computational Linguistics. pp. 5268--5277 (2019)

\bibitem{fan2019target}
Fan, Z., Wu, Z., Dai, X., Huang, S., Chen, J.: Target-oriented opinion words
  extraction with target-fused neural sequence labeling. In: Proceedings of the
  2019 Conference of the North American Chapter of the Association for
  Computational Linguistics: Human Language Technologies, Volume 1 (Long and
  Short Papers). pp. 2509--2518 (2019)

\bibitem{HellBen18}
Hellstr{\"{o}}m, T., Bensch, S.: Understandable robots. Paladyn  \textbf{9}(1),
   110--123 (2018)

\bibitem{honnibal2017spacy}
Honnibal, M., Montani, I.: spacy 2: Natural language understanding with bloom
  embeddings, convolutional neural networks and incremental parsing. To appear
  \textbf{7} (2017)

\bibitem{peng2019knowing}
Peng, H., Xu, L., Bing, L., Huang, F., Lu, W., Si, L.: Knowing what, how and
  why: A near complete solution for aspect-based sentiment analysis. arXiv
  preprint arXiv:1911.01616  (2019)

\bibitem{perkins2014python}
Perkins, J.: Python 3 text processing with {NLTK} 3 cookbook. Packt Publishing
  Ltd (2014)

\bibitem{pontiki2016semeval}
Pontiki, M., Galanis, D., Papageorgiou, H., Androutsopoulos, I., Manandhar, S.,
  Mohammad, A.S., Al-Ayyoub, M., Zhao, Y., Qin, B., De~Clercq, O., et~al.:
  Semeval-2016 task 5: Aspect based sentiment analysis. In: Proceedings of the
  10th international workshop on semantic evaluation (SemEval-2016). pp. 19--30
  (2016)

\bibitem{socher2013recursive}
Socher, R., Perelygin, A., Wu, J., Chuang, J., Manning, C.D., Ng, A., Potts,
  C.: Recursive deep models for semantic compositionality over a sentiment
  treebank. In: Proceedings of the 2013 conference on empirical methods in
  natural language processing. pp. 1631--1642 (2013)

\bibitem{sutherlandleveraging}
Sutherland, A., Magg, S., Wermter, S.: Leveraging recursive processing for
  neural-symbolic affect-target associations. In: 2019 International Joint
  Conference on Neural Networks (IJCNN). pp.~1--6. IEEE (2019)

\bibitem{tai2015improved}
Tai, K.S., Socher, R., Manning, C.D.: Improved semantic representations from
  tree-structured long short-term memory networks. In: Proceedings of the 53rd
  Annual Meeting of the Association for Computational Linguistics and the 7th
  International Joint Conference on Natural Language Processing (Volume 1: Long
  Papers). pp. 1556--1566 (2015)

\bibitem{wang2016recursive}
Wang, W., Pan, S.J., Dahlmeier, D., Xiao, X.: Recursive neural conditional
  random fields for aspect-based sentiment analysis. In: 2016 Conference on
  Empirical Methods in Natural Language Processing. pp. 616--626 (2016)

\bibitem{wu2016google}
Wu, Y., Schuster, M., Chen, Z., Le, Q.V., Norouzi, M., Macherey, W., Krikun,
  M., Cao, Y., Gao, Q., Macherey, K., et~al.: Google's neural machine
  translation system: Bridging the gap between human and machine translation.
  arXiv preprint arXiv:1609.08144  (2016)

\end{thebibliography}

\end{document}